\documentclass{article}
\PassOptionsToPackage{numbers,sort&compress}{natbib}
\usepackage[T1]{fontenc}
\usepackage{arxiv}
\renewcommand{\shorttitle}{STAMP: Generative Design and Shape Optimization}
\renewcommand{\undertitle}{}
\renewcommand{\headeright}{}
\usepackage{etoolbox}
\makeatletter
\patchcmd{\@maketitle}{\textsc{\undertitle}\\}{}{}{\PackageError{STAMP}{Title label removal failed}{}}
\makeatother
\usepackage{amsmath,amssymb,graphicx,booktabs,tabularx,array,microtype,float,placeins,natbib,longtable}
\usepackage[hidelinks]{hyperref}
\newcolumntype{Y}{>{\raggedright\arraybackslash}X}

\hypersetup{pdftitle={STAMP: Manufacturing-Informed Generative Design and Shape Optimization of Sheet Panels}}
\hypersetup{pdfauthor={Jihoon Kim, Jun Hyeok Song, Wonho Ha, Namwoo Kang}}
\title{STAMP: Manufacturing-Informed Generative Design and Shape Optimization of Sheet Panels}
\author{\normalfont Jihoon Kim$^{1,2}$\quad Jun Hyeok Song$^{2}$\quad Wonho Ha$^{2}$\quad Namwoo Kang$^{1,2}$\thanks{Corresponding author: nwkang@kaist.ac.kr} \\ 
{\small\normalfont $^{1}$ Department of Mechanical Engineering} \\ 
{\small\normalfont Korea Advanced Institute of Science and Technology, Daejeon 34141, Republic of Korea} \\ 
{\small\normalfont $^{2}$ Narnia Labs, Daejeon 34051, Republic of Korea}}
\date{}
\begin{document}
\maketitle
\begin{abstract}

Generative design of formed sheet panels must preserve geometric manufacturing conditions while exploring layouts and improving performance. STAMP embeds these conditions directly in the shape representation. Each panel is described by its material layout and a reference surface with one height at each occupied position in a plane perpendicular to the drawing direction. Nominal thickness is prescribed along the normal. The representation allows curved surfaces, ribs, and openings. It excludes reference-surface undercuts and keeps nominal thickness fixed without manufacturing penalty terms. Because these conditions belong to the representation, every generated and optimized panel satisfies them regardless of training-set size. Boundary-conditioned diffusion generates alternative layouts, and predictor gradients guide a surface optimization that retains outlines, openings, and designated fixed regions. We demonstrate STAMP on automotive hood inner panels using CarHoods10k. With a comparable number of stored values, depth maps reconstruct surfaces more accurately than point-cloud and voxel baselines. From only five training designs, one of three training runs yields novel and varied layouts. A generator trained on 500 designs, 6.5\% of the training set, already reaches the layout variety of the full-data model. Across 11 held-out design families, predictor-based selection among existing dataset designs yields an average gap of 0.18 mm to the best available FEA deformation, versus an expected 1.59 mm for random selection. It does better than random selection in all 11 families. At a 10\% depth-change bound, a network not used in the search predicts an average deformation reduction of 24.8\% for six generated and existing panels, with a 3.2\% average mass increase computed from geometry. Engineers can thus explore and optimize panels within this manufacturing-informed design space.

\end{abstract}
\section{Introduction}
Industrial generative design must connect geometric exploration with manufacturing requirements and performance-driven modification. Deep generative models expand the range of design alternatives \cite{regenwetter2022} and can couple generation with structural objectives and diversity goals \cite{oh2019,padgan2021}. CAD/CAE integration further links generated concepts to performance evaluation \cite{kang2025scenarios,yoo2021cadcae}. Putting these capabilities into practice also depends on data availability. Obtaining varied engineering geometries with corresponding analysis results remains a challenge, motivating synthetic datasets and methods that learn from limited examples \cite{hong2025deepjeb,kwon2025limited}.

Manufacturing knowledge can help structure this design problem. Design-for-manufacturing methods incorporate process requirements into component design \cite{kuo2001,patterson2021}, while manufacturing-oriented representations can embed geometric requirements in the design variables themselves \cite{herholz2015,paireddepth}. Machine learning has also been applied to sheet-forming process simulation and optimization \cite{cao2024}. At the component-design level, the challenge is to preserve geometric manufacturing conditions and prescribed interfaces while retaining the freedom to create and improve alternative panel geometries.

We describe panel geometry by its surface height along the drawing direction, the direction in which the forming tool moves into the sheet, and by the region occupied by material. At each occupied position in the perpendicular plane, the reference surface has one height (a single-valued reference surface). Nominal thickness is prescribed separately along the surface normal. By construction, the resulting design space excludes reference-surface undercuts along the drawing direction and maintains the prescribed nominal thickness, so it needs no manufacturing penalty terms for these conditions. The design space still admits curved profiles, ribs, varied outlines, and opening layouts without a predefined catalogue of feature templates. In the continuous setting, height and material region fully specify such a reference surface.

We present STAMP, a manufacturing-informed representation and generative design framework that uses this design space for layout exploration and performance-driven shape optimization (Fig. 1). Prescribed outer domains and boundary heights condition the generation of alternative interior layouts. Structural predictions support candidate selection, and predictor gradients guide B-spline free-form deformation (FFD) of a chosen layout. The deformation retains the layout's outline, openings, and designated fixed regions. Engineers can thus explore alternative layouts and improve predicted structural performance while preserving the encoded geometric manufacturing conditions and specified interfaces. To the best of our knowledge, STAMP is the first framework in which boundary-conditioned generation, structural performance prediction, and gradient-based shape optimization of formed sheet panels all operate directly on a shared depth map that embeds these geometric manufacturing conditions in the design variables.

\begin{figure}[tp]
\centering
\includegraphics[width=\linewidth,height=205mm,keepaspectratio]{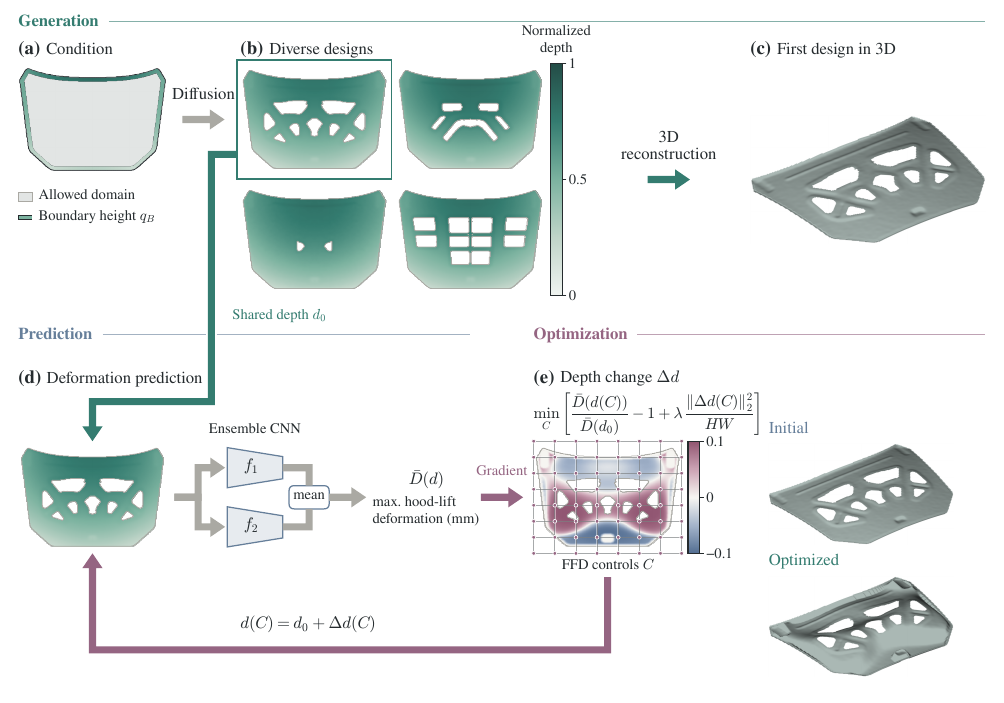}
\caption{STAMP design framework, illustrated with a CarHoods10k case study. Panels show (a) the prescribed boundary, (b) generated depth maps, (c) the reconstructed surface, (d) ensemble deformation prediction, and (e) FFD optimization. The overlaid grid shows the projected FFD controls $C$, and $\bar D$ denotes the mean predicted deformation. Depths are normalized. Examples come from three full-data training runs ($N$ = 7,660 training designs). Optimization uses a 10\% depth-change bound and preserves the outline, openings, prescribed boundary heights, and hinge and latch mounts. Depth changes in the optimized 3D view are magnified 2$\times$ for clarity.}\label{fig:figure1}
\end{figure}

We use automotive hood inner panels as a case study for this representation and design framework. A hood inner panel must join a prescribed surrounding interface while allowing alternative surface profiles and opening layouts. Shape-grammar and data-driven methods support automated hood exploration \cite{mccormack2002,carhoods2022}. The design task goes beyond proposing a layout to improving its structural performance within that interface. We evaluate STAMP on CarHoods10k \cite{carhoods2022,dataset}, training and testing the generator and structural predictors on this hood design problem.

Experiments show lower surface reconstruction error for the depth map than for the point-cloud and voxel baselines at comparable numbers of stored values. Conditional generation produces alternative panel layouts under a shared boundary. From only five training designs, one of three training runs yields novel and varied layouts, and a generator trained on 500 designs, 6.5\% of the training set, already reaches the layout variety of the full-data model. On held-out design families, predictor-based selection reduces the gap to the lowest finite-element analysis (FEA) deformation among the available designs by 89\% relative to the expected gap under uniform random selection, and it does better than random choice in all 11 families. Gradient-based optimization reduces predicted deformation in generated and existing panels while retaining their prescribed boundary heights, openings, and designated fixed regions, with mass changes computed from geometry. In the hood case study, these results demonstrate geometric fidelity, layout exploration, and performance-driven surface development within the manufacturing-informed panel design space.

\section{Related work}
\subsection{Manufacturing-informed shape representations}
Manufacturing-informed design methods encode process knowledge in the set of geometries that can be created or modified. The representation determines which features are easy to express and which conditions can be retained during exploration. Design-for-manufacturing research spans design simplification and the generation and enforcement of process-specific constraints \cite{kuo2001,patterson2021}. Manufacturing-oriented topology optimization enforces such constraints on a density field, for example by restricting material changes to a casting or stamping direction \cite{gersborg2011,liu2016survey,yuan2024stamping}. The choice of representation matters for generative design because process knowledge can then determine the design variables themselves rather than serve only as a check on a completed shape.

For sheet-metal parts made by cutting and bending, manufacturing knowledge can be expressed through discrete features and assembly rules. Barda et al. \cite{barda2023} represent load-bearing fixtures through a grammar of planar tabs and bends, combining stochastic design search, continuous geometry modification, and finite-element evaluation. SheetGen \cite{sheetgen2026} enumerates connections between planar mounting tabs using explicit bending rules and evaluates the resulting parts through CAD reconstruction and CAM bending simulation. These methods organize folded parts around process-related features. STAMP addresses smoothly formed panels through a continuous surface field and an explicit material region.

Surface and implicit representations provide further ways to describe geometry for manufacturing. Herholz et al. \cite{herholz2015} partition and deform free-form meshes into direction-dependent height-field patches for milling and mould fabrication. Kim et al. \cite{paireddepth} use two-dimensional depth images to impose die-casting and injection-moulding requirements in generative design. Their method removes non-manufacturable overhangs and produces designs with a two-dimensional generative model. Implicit neural representations describe shapes through continuous signed distance functions (SDFs) \cite{deepsdf2019}, and stamping-specific formulations have improved their ability to represent small geometric features \cite{attarimplicit2023}. These approaches address geometric decomposition, cast and moulded solids, or implicit shape modelling. STAMP represents the panel reference surface directly on a two-dimensional grid. The same field encodes the footprint and openings, and nominal thickness is prescribed separately. This panel-specific description gives direct access to the manufacturing-informed geometry without an intervening volumetric decoder and provides shared design variables for generation, performance assessment, and shape optimization.

\subsection{Conditional generative design of sheet panels}
Deep generative design combines learned geometric variation with the conditions that define an engineering problem \cite{regenwetter2022}. Oh et al. \cite{oh2019} integrate generative models and topology optimization to expand a collection of structural designs. Jang et al. \cite{jang2022diversity} use reinforcement learning to select topology optimization parameters that promote design diversity. TopologyGAN \cite{topologygan2021} conditions structural layout generation on physical fields computed over the initial material domain. PaDGAN \cite{padgan2021} promotes both design quality and diversity, while RANGE-GAN \cite{rangegan2022} generates alternatives under specified attribute ranges. These studies address complementary parts of design exploration. For formed panels, the condition must also describe the geometric interface that the generated interior joins.

Diffusion models offer a flexible approach to conditional design generation \cite{ddpm2020,leng2025}. TopoDiff \cite{topodiff2023} guides topology generation with performance and manufacturability surrogates, while Diffusion Optimization Models \cite{dom2023} align denoising trajectories with physics-based optimization trajectories. Herron et al. \cite{herron2024} generate three-dimensional voxel structures with latent diffusion conditioned on initial strain-energy fields and use a convolutional surrogate to evaluate generated candidates. Zhang et al. \cite{zhang2025} combine latent diffusion with solid isotropic material with penalization (SIMP) in multistage topology optimization and apply the approach to automotive hoods. These methods explore material occupancy and structural topology. STAMP instead generates the height and opening layout of a single-valued sheet panel under a prescribed geometric interface, and then optimizes its surface profile in the same explicit representation.

Image completion provides a related view of the interface condition. RePaint \cite{repaint2022} uses an unconditional diffusion model to fill unknown image regions while incorporating known pixels through noise-matched sampling and resampling. STAMP trains a conditional model on panel geometry and restores the prescribed field values during sampling. The boundary therefore serves as a fixed engineering interface, while the interior remains available for alternative layouts.

Automotive hood generation gives this conditioning problem a concrete setting. McCormack and Cagan \cite{mccormack2002} synthesize hood inner panels with a parametric shape grammar. Sugiura et al. \cite{sugiura2026} use a conditional variational autoencoder with B-spline boundary curves, representing the panel interior by a material-presence flag and a height residual relative to a Coons patch. Their formulation connects a supplied boundary to a reference-surface description. They report that generating heights directly, without this residual, left large deviations from the supplied boundary, and that the residual reduced but did not eliminate them. STAMP generates panel depth directly and imposes the known geometric values at every diffusion sampling step. It also passes the resulting field to structural prediction and FFD optimization. This connects exploration of several interior layouts for one interface to the performance-driven development of a selected panel.

\subsection{Performance prediction and surrogate-guided design optimization}
Learned performance models allow many design alternatives to be evaluated at the cost of a model prediction. Surrogate-based analysis and optimization establish the broader connection between approximate response models, sensitivity analysis, and design search \cite{queipo2005}. Convolutional models have predicted structural fields directly from image-based geometry and loading information \cite{nie2020stress}. Datasets such as DeepJEB pair generated geometries with structural analysis results to support performance prediction \cite{hong2025deepjeb}. In forming, geometry-based deep learning also supports rapid prediction of thinning and wrinkling \cite{attarrapid2021,cao2024}. These applications show why the representation supplied to a predictor should retain the geometric information needed for the target response.

For sheet stamping, Attar et al. \cite{attar2023} couple an SDF geometry decoder with a height-map CNN and update the latent code to satisfy predicted hot-stamping thinning constraints. Zhou et al. \cite{zhou2025} optimize the blank outline for a hot-stamped B-pillar using a two-dimensional SDF auto-decoder, a thickness-field CNN, and latent-code updates. These searches modify stamping geometry or the undeformed blank for forming-process objectives. Attar et al. \cite{attar2023} identify in-service measures such as stiffness or crashworthiness as a direction for extending their platform. STAMP evaluates structural deformation under a hood-lift load case and applies predictor gradients directly to B-spline FFD control values, retaining the selected panel outline and opening pattern.

CarHoods10k connects automotive hood geometry, structural analysis labels, learned representations, and design optimization \cite{carhoods2022}. With its point-cloud representation and latent-space search, it poses a public problem in performance-driven panel design. Its performance values have been predicted from point-cloud latent codes \cite{carhoods2022} and, since then, from images combined with cross-sections \cite{indupally2025mmml} and from attributed feature graphs \cite{indupally2026afg}. Beyond this dataset, Rostami et al. \cite{rostami2025hood} reduce the design variables of a hood inner panel through a two-dimensional mapping and optimize its shape and size for head injury, eigenfrequency, and static load using machine-learning approximations. A recent preprint by Chavare \cite{chavare2026} combines topology-family retrieval, point-cloud deformation conditioned on key performance indicators, and a neural-operator surrogate to assess candidate hood inner panels. In that work, the requested condition specifies performance attributes and a design family. STAMP instead takes a geometric boundary condition, generates interior alternatives as depth maps, and optimizes the surface of a selected layout.

Beyond sheet panels, learned representations and constrained deformation support shape exploration. Rios et al. use the latent space of a point-cloud autoencoder as a unified representation for multitask shape optimization of car bodies \cite{rios2022multitask}, and Point2FFD decodes such latent codes into simulation-ready meshes by deforming templates \cite{rios2021point2ffd}. In these methods, the search moves through a learned latent space, whereas STAMP edits the explicit panel field, so prescribed boundary heights and openings stay fixed during optimization. Kwon and Kang \cite{kwon2025limited} learn an implicit shape representation from limited datasets and optimize latent codes with a genetic algorithm, evaluating candidates through structural and flow simulations. Padula et al. \cite{padula2024} combine constrained FFD with latent generative models and a coordinate-correction layer for geometric constraints, evaluating generated shape distributions with Poisson and naval-flow simulations. FFD produces smooth shape changes through a control lattice \cite{ffd1986}. In STAMP, FFD is applied after generation, and structural predictor gradients drive it to optimize the panel depth map directly while preserving its prescribed regions.

\section{Methods}

STAMP embeds geometric manufacturing requirements in a design space defined by a single-valued reference surface, its material region, and a separately prescribed nominal thickness (Fig. 1). By construction, the reference surface satisfies the drawing-direction condition and the panel keeps its nominal thickness. The depth map also carries prescribed boundary heights and supplies directly editable geometry for conditional generation, structural prediction, and gradient-based optimization.

\subsection{Manufacturing-informed representation}
We consider formed sheet panels whose reference surface has a single height at each occupied location in a selected projection plane. Let $\Omega$ denote the region occupied by material, with openings excluded. The surface is

\begin{equation}
\mathbf{s}(x,y)=(x,y,h(x,y)),\qquad (x,y)\in\Omega.
\end{equation}

The choice of projection direction defines the admissible shape family. The single-height condition excludes undercuts of the reference surface along this direction, places where a line parallel to this direction meets the reference surface more than once. It expresses a geometric restriction motivated by directional tool access, as in casting- and stamping-direction constraints for topology optimization \cite{gersborg2011,liu2016survey,yuan2024stamping}. Curved panels, ribs, and openings can be represented provided that the reference surface remains single-valued in this direction. Nominal thickness is specified separately along the surface normal. Thus, changes in surface height describe panel shape, while thickness remains a prescribed design attribute. This formulation makes the geometry directly editable on a two-dimensional grid (Fig. 2).

We encode material height and openings in one depth map. For a common height range $[h_{\min},h_{\max}]$, the normalized material depth is $d=(h-h_{\min})/(h_{\max}-h_{\min})$. The stored field is

\begin{equation}
q_{ij}=\begin{cases}d_{ij}, & (x_i,y_j)\in\Omega,\\-1, & (x_i,y_j)\notin\Omega.\end{cases}
\end{equation}

Material values lie in $[0,1]$, and the material mask is recovered by thresholding the field at $-0.5$. The mask describes where material exists and does not introduce a second surface. The same field specifies the geometry used by generation, prediction, and optimization. A surface mesh is obtained by restoring the height scale and connecting adjacent material pixels without bridging openings.

The representation encodes geometric manufacturing requirements in its admissible variables. Any height function defines a single-valued reference surface in the selected direction, and nominal thickness remains a prescribed constant along the surface normal. These conditions require no manufacturing penalty terms. Boundary heights are explicitly imposed during generation and fixed during shape optimization. Designated hinge and latch regions are additionally fixed during optimization, preserving their initial geometry. Where symmetry is required, it is imposed by reflection and by symmetric optimization increments. In the continuous setting, height and material region fully specify the reference surface within this shape class. Sampling and quantization on the finite grid introduce approximation errors, which we assess in Section 4.1.

\begin{figure}[tp]
\centering
\includegraphics[width=\linewidth,height=205mm,keepaspectratio]{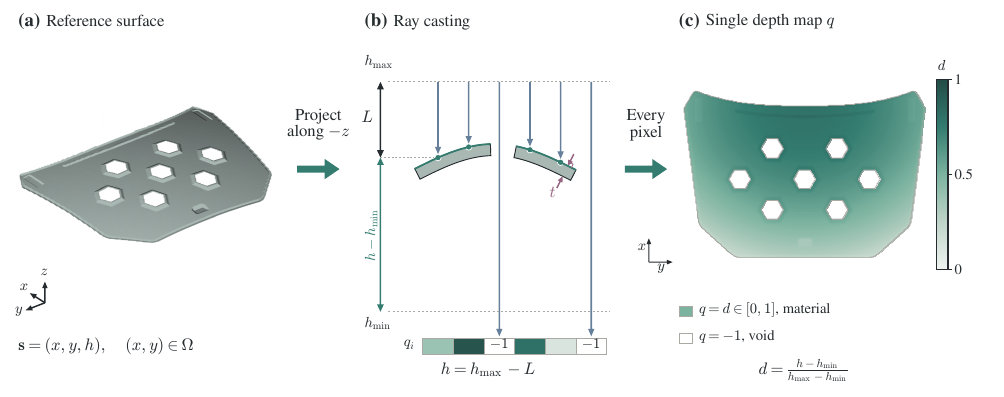}
\caption{Single-depth representation of a formed sheet panel, illustrated with a CarHoods10k example. (a) Reference surface. (b) Schematic orthographic rays give the first-hit distance $L$, which is converted to the normalized height $d$ and stored per pixel. Openings and the exterior take $q=-1$. Nominal thickness $t$ is prescribed along the normal. Colours in (b) span $d$ = 0.64--0.72. (c) The resulting depth map of the surface in (a).}\label{fig:figure2}
\end{figure}

\subsection{Data and design conditions}
We evaluate STAMP on CarHoods10k~\cite{dataset}, a public dataset of hood-frame geometries with structural performance values for a hood-lift load case. The experiments use a derived release in normalized OBJ format, matched to the supplied performance labels by design identifier. Of the 10,070 label rows in the release, 9,644 have a corresponding derived mesh. These designs come from 104 base geometries, which the dataset files label as skins. Each base geometry combines one of ten hood surfaces with one of eleven pocket-feature patterns, and its designs vary the feature parameters \cite{carhoods2022}. We call each base geometry a design family. Families are assigned to disjoint training, validation, and test partitions, containing 7,660, 912, and 1,072 designs from 83, 10, and 11 families, respectively (Table 1). The label targets are maximum equivalent stress, mass, and maximum directional deformation.

The dataset applies 150 N loads at two front corners and remote-displacement supports at the two hinge regions and the hood latch \cite{carhoods2022}. The deformation target, the maximum displacement in the z direction in millimetres, characterizes the panel response to this hood-lift loading. In nine families, the dataset also varies hinge and latch parameters between designs. These parameters are not visible in the geometry, so part of the label variation in those families is not represented in the input.

Each mesh is projected onto a $256\times256$ grid in a shared coordinate frame. The frame is calibrated from one design in each of the 83 training families. The projection direction is the axis with the smallest median geometric extent, which is the $z$ direction for these data and which we take as the drawing direction. Orthographic rays cast from $h_{\max}$ along $-z$ sample the first surface intersection at the grid's pixel centres (Fig. 2b). A hit at distance $L$ gives $h=h_{\max}-L$. Rays without an intersection identify openings or the exterior. All designs use the same planar bounds and height normalization. Reconstruction returns the visible reference surface in the normalized OBJ coordinates. The official release stores each design in millimetres as a closed sheet with a nearly uniform thickness of 1.20 mm, which matches the prescribed-thickness formulation. The derived release reproduces these meshes triangle for triangle and scales each design uniformly so that its longest side spans 1.818 units. One unit therefore corresponds to 843 to 975 mm, depending on the family.

The generation input specifies the panel's outer domain and its boundary height. We obtain the domain $D$ by filling the openings of the largest connected material region. A three-pixel boundary band defines the fixed region $B$, and the stored values on that band define the known values $q_B$. The condition consists of three channels $q_B$, $B$, and $D$. Interior openings are left unspecified so that the generator can propose alternative layouts within the same boundary. These are geometric design conditions, distinct from the loads and supports underlying the structural labels.

\begin{table}[tp]
\centering
\caption{Dataset partitions. Families remain disjoint across partitions.}\label{tab:table1}
\begin{tabular}{lrr}
\toprule
Partition & Designs & Families \\
\midrule
Training & 7,660 & 83 \\
Validation & 912 & 10 \\
Test & 1,072 & 11 \\
Total & 9,644 & 104 \\
\bottomrule
\end{tabular}
\end{table}

\subsection{Conditional generation}
Following the denoising diffusion formulation \cite{ddpm2020}, a conditional diffusion model generates the unknown interior while retaining the prescribed boundary. Its denoising network, a two-dimensional U-Net \cite{unet2015}, receives the noisy depth map together with the three condition channels and predicts one noise field.

Let $U=D\odot(1-B)$ be the editable region, with $\odot$ denoting elementwise multiplication, and let $k$ be the field containing $q_B$ on the boundary and $-1$ outside the domain. At each training step, a randomly selected diffusion timestep and Gaussian noise produce a noisy sample $q_t$. Before denoising, the known values are restored as

\begin{equation}
\widetilde q_t=U\odot q_t+(1-U)\odot k.
\end{equation}

Training minimizes the squared noise-prediction error over editable pixels. For a minibatch, the loss is

\begin{equation}
\mathcal L=\frac{\sum_{b,i,j}U_{bij}\bigl(\epsilon_{\theta}(\widetilde q_t,c,t)_{bij}-\epsilon_{bij}\bigr)^2}{\sum_{b,i,j}U_{bij}},
\end{equation}

where $c$ denotes the condition channels. At inference, denoising diffusion implicit model (DDIM) sampling \cite{ddim2021} starts from Gaussian noise and restores the known values after every reverse step. The reported sampling configuration uses 50 steps with $\eta=0$. Different initial noise samples produce alternative designs under the same geometric input.

Decoding separates the generated material region from its height values. A mask-normalized Gaussian filter with a standard deviation of two pixels smooths material height without averaging in the empty region. The operation preserves the generated opening mask, clips material depth to $[0,1]$, and restores the fixed boundary heights. We keep the unsmoothed and smoothed outputs separately for evaluation.

When a design requires bilateral symmetry, we reflect the depth and material pattern of the negative-$y$ half onto the positive-$y$ half. This explicit design constraint is applied after generation to inputs selected for symmetry-compatible outer domains and boundary heights. The same rule is used across the data-size comparison. Boundary agreement is checked after reflection because the prescribed boundary heights can be slightly asymmetric. The resulting depth map can be reconstructed as a surface, evaluated by the predictor, and used as the starting point for shape optimization.

Additional training settings, hardware specifications, and computational costs are provided in the supplementary material.

\subsection{Depth-based performance prediction}
The same depth representation supports both design generation and structural performance prediction. A convolutional neural network (CNN) maps the normalized depth map and material mask to stress, mass, and deformation under the CarHoods10k hood-lift loading condition. Deformation drives candidate selection and shape optimization, and mass is predicted for comparison across designs. Within an optimization, mass changes are computed directly from the surface area at the prescribed nominal thickness. Stress is included in joint training but is not used in the optimization objective, constraints, or candidate selection. Its test errors are reported in Supplementary Section S6. Masked coordinate channels encode the position of geometric features within the common design frame. This allows generated panels to be evaluated directly in the representation used to construct them.

The predictor uses convolutional feature extraction and a fully connected regression head. Stress and deformation are transformed logarithmically, and all targets are standardized using training statistics. Predictions are converted back to their physical units before the outputs of the three independently trained models are averaged.

\subsection{Gradient-based shape optimization}
STAMP connects generated panels to performance-driven shape optimization in the same depth representation. The optimizer adjusts surface height while retaining the panel outline, opening pattern, designated hinge and latch regions, and prescribed geometric conditions. An FFD control lattice \cite{ffd1986} provides smooth shape changes. Our cubic B-spline field maps an 8 $\times$ 8 control grid to the depth map. Gradients pass through the predictor and deformation field to update the control grid, allowing the search to operate directly on the generated design.

Let $d_0$ be the initial normalized depth map and $\Delta d(C)$ the increment defined by control values $C$. We parameterize the candidate as

\begin{equation}
d(C)=d_0+\Delta d(C), \qquad
\Delta d(C)=r\,E\odot\tanh(F(C)).
\end{equation}

Here, $F$ is the symmetrized B-spline field, $r$ bounds the depth change relative to the common depth range, and $E$ combines the movable region, boundary taper, and depth envelope. The depth-change bound is selected for the design problem rather than imposed by the representation. The increment vanishes in the prescribed fixed regions and transitions smoothly into the editable interior. The fixed regions include the boundary band and the designated hinge and latch regions. Material around internal openings can move with the surrounding surface while the opening mask remains fixed. The increment is symmetric and keeps depths within the allowed range. Any existing asymmetry in the initial surface remains. Supplementary Section S6 gives the construction of $E$. The uniform bound $r$ is the simplest pointwise envelope, and spatially varying lower and upper height limits can take its place in the same parameterization.

Two fixed CNNs provide the optimization objective. Each prediction is converted back to deformation in millimetres before averaging. With their mean denoted by $\bar D$, we minimize

\begin{equation}
J(C)=\frac{\bar D(d(C))}{\bar D(d_0)}-1
+\frac{\lambda}{HW}\sum_{i=1}^{H}\sum_{j=1}^{W}\Delta d_{ij}(C)^2,
\qquad \lambda=1.
\end{equation}

The quadratic term penalizes changes from the initial surface across the full $H\times W$ image grid, with $H=W=256$. Mass is monitored, but not constrained, to show the trade-off that accompanies deformation reduction. Nominal thickness is not an optimization variable.

\subsection{Evaluation protocol}

\paragraph{Generation} The full-data generator uses all 7,660 training designs. To assess sensitivity to data availability, we also train with 5, 10, 50, 100, and 500 designs, and we denote the number of training designs by $N$. For each of three training runs, R1--R3, the smaller sets are nested prefixes of a randomly permuted training pool sampled without replacement, so each smaller set is contained in the next larger one. The runs vary both the training seed and the selected subset. Validation and test partitions remain fixed.

All conditions use the same network, image resolution, batch size of 32, and AdamW \cite{adamw2019} learning rate of $10^{-4}$. Each training cycle presents 7,660 examples. Smaller datasets repeat shuffled passes to fill this budget, giving 240 optimizer updates per cycle. The reported comparison uses the final checkpoint after 1,000 cycles rather than each model's most favourable checkpoint. This holds optimization effort constant while varying the number of available designs. Supplementary Table S2 lists the training costs. Validation loss measures noise-prediction error and is used to monitor training rather than to measure shape quality directly.

Generated shapes are assessed with fixed validation inputs and repeated noise samples. These validation inputs share their outer domains and boundary heights with training families that use other feature patterns, so the comparison assesses alternative layouts for a known panel boundary rather than for a new hood surface. The main data-size comparison uses eight samples for each of two symmetry-compatible inputs, Input A and Input B, per trained model. The two metrics defined next are computed on the decoded fields after height smoothing and reflection (Section 3.3). Surface roughness is the root mean square height residual relative to mask-normalized Gaussian smoothing ($\sigma=1$ pixel), excluding a three-pixel material-edge band. Pattern diversity is the mean Jaccard distance, $1-|M_a\cap M_b|/|M_a\cup M_b|$, over the 28 pairs of generated material masks. Lower roughness indicates smaller fine-scale height fluctuations. A higher Jaccard distance indicates more varied opening and material layouts. Values are averaged equally over the two inputs. The plots show the mean and the observed range across the three training runs.

A supplementary training-budget study trains three new five-design models from scratch with the same subsets and seeds, evaluating each at preset milestones from 500 to 10,000 updates. The comparison uses the same two inputs, sampling, smoothing, and reflection as the data-size evaluation. The original 240,000-update models provide the reference (Supplementary Figs. S3 and S4).

\paragraph{Prediction} The predictor uses 7,660 training designs, 912 validation designs, and 1,072 test designs. No design family appears in more than one partition. The test set contains 11 held-out families, labelled F1--F11. Each held-out family shares its outer domain and boundary heights with training families that use other feature patterns. Three networks are trained for 200 epochs using AdamW \cite{adamw2019} with a learning rate of $10^{-4}$, weight decay of 0.01, and batch size 64. Batches include groups of eight designs from the same design family. The training loss combines standardized target error with an additional error term on differences between designs from the same family. The latter has weight 2. Checkpoints are selected using validation deformation error and within-family differences, with the differences scaled by their variation in the training set. The selected checkpoints remain fixed for test evaluation and optimization.

We assess absolute prediction accuracy using the mean absolute error (MAE), root mean square error (RMSE), and $R^2$. We also evaluate the design decisions supported by the predictor. Within each design family, pairwise ranking accuracy measures how often the predictor identifies which design in a pair has the lower FEA deformation. Pairs tied in FEA deformation are excluded, and tied predictions receive half credit. Selection error is the FEA deformation of the candidate with the lowest predicted deformation minus the best available FEA deformation. We compare this error with the expected error of a uniformly random candidate. Tied predicted minima are averaged. Candidate-selection errors are averaged equally across design families. We also compare the FEA deformation of the selected and best available candidates with the expected FEA deformation under uniform random selection. During pilot development, test metrics were also computed when comparing $128\times128$ with $256\times256$ inputs and when testing a ResNet50 backbone. The final networks use $256\times256$ inputs and the architecture in Section 3.4, and their training settings and checkpoints were selected on the validation partition only.

\paragraph{Optimization} We optimize three generated panels (G1, G2, G3) and three panels from the dataset (O1--O3). G1, G2, O1, and O2 share Input A. G3 and O3 use Input B. The six designs were selected for symmetry before any optimization results were inspected. O1--O3 and Inputs A and B come from validation families, which were also used for checkpoint selection. We optimize every starting design at each of four depth-change bounds, 5\%, 10\%, 15\%, and 20\%, chosen for this comparison.

The two optimization CNNs and a third evaluation CNN use the fixed predictor checkpoints selected under Prediction above. The third CNN contributes neither gradients nor the choice of the final state. Adam \cite{adam2015} updates the control grid, initialized to zero, for 200 updates with a learning rate of 0.05. We keep the state with the lowest full objective as the final state, including the initial state in the comparison. The same starting designs, checkpoints, and update budget are used at all four bounds. Supplementary Table S1 reports the predicted deformation reductions and area-based mass changes for all 24 searches.

All improvements reported for optimized shapes are predictor estimates under the learned hood-lift response, rather than new FEA results.

\section{Results}

\subsection{Representation accuracy}
The 256 $\times$ 256 depth map provides the geometric input shared by all three design stages. We evaluated its reconstruction on one fixed design from each of the 104 dataset families. Distances were measured from visible surface points sampled on a 512 $\times$ 512 reference grid to the reconstructed triangles. In the common normalized coordinates, the median mean distance across designs was $0.142\times10^{-3}$ (0.13 mm) and the median 95th-percentile distance was $0.753\times10^{-3}$ (0.68 mm). The mean distance is about a tenth of the 1.20 mm sheet thickness. Supplementary Fig. S1 reports the complete distribution and the resolution comparison. These measurements quantify the surface approximation used by the framework at its working resolution.

Other shape representations reproduce the panel less closely when they store the same number of values (Fig. 3). We compared the depth map with a point cloud and a voxel occupancy grid, each limited to about 65,536 stored values and evaluated against the same reference points (Supplementary Section S1). A voxel grid spends most of its cells on empty space and records only whether each cell is occupied. At this budget its cells are three to four times wider than a depth-map pixel, and its median mean distance is $9.51\times10^{-3}$, 67 times that of the depth map. A point cloud stores two in-plane coordinates with every height, so the same budget holds 21,845 points. Triangulated in the projection plane with the most favourable edge-length limit for each design, these points give a median mean distance of $0.210\times10^{-3}$, 1.5 times that of the depth map. Openings must then be inferred from gaps between points, and the opening region is recovered with a median intersection over union (IoU) of 0.82, compared with 0.94 for the depth map and 0.70 for the voxel grid (Supplementary Table S3). Of the three representations, the depth map gives the smallest distance and the highest opening overlap for all 104 designs. This advantage comes from the same condition that excludes reference-surface undercuts. The depth-map grid fixes the in-plane positions, so every stored value records height, whereas the point cloud spends two thirds of its values on in-plane coordinates and most voxels record empty space.

\begin{figure}[tp]
\centering
\includegraphics[width=\linewidth,height=205mm,keepaspectratio]{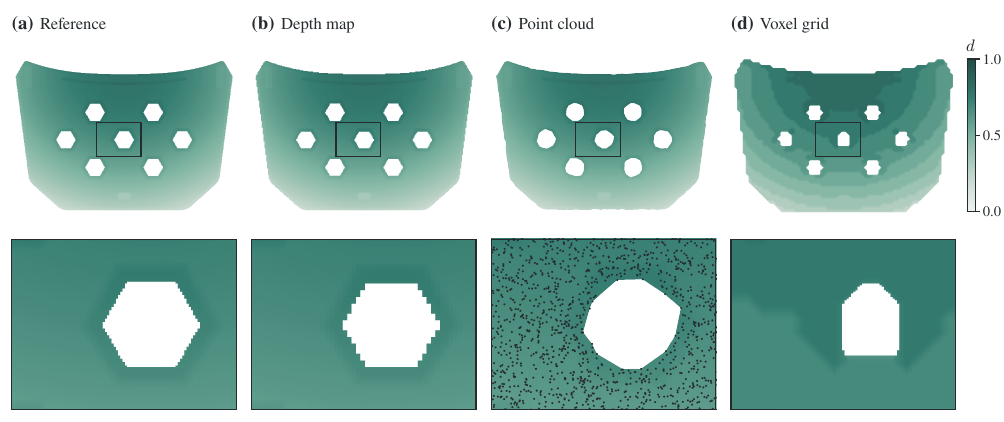}
\caption{Representations of the panel in Fig. 2. (a) Reference surface sampled on a 512 $\times$ 512 grid. (b--d) Representations that each store about 65,536 values. (b) 256 $\times$ 256 depth map. (c) Surface triangulated from 21,845 points, with the stored points marked in the enlarged view. (d) 60 $\times$ 78 $\times$ 14 voxel occupancy grid. The lower row enlarges the boxed region five times.}\label{fig:figure3}
\end{figure}

\subsection{Conditional generation and data availability}
STAMP generates alternative panel interiors under the same prescribed geometric boundary (Fig. 4). The two illustrated inputs each produce four candidates with different material and opening layouts. Their depth maps convert directly into three-dimensional reference surfaces, so the generated image is also the design geometry used by the subsequent predictor and optimizer.

The symmetry-constrained outputs stay within the panel's outer domain. Because the known values are restored at every sampling step and the selected inputs have compatible outer domains, no generated material lies outside the prescribed outer domain after reflection in the two-input data-size evaluation. The largest boundary-height discrepancy after reflection is $7.3\times10^{-4}$ of the common depth range, which arises from a slight asymmetry in the prescribed boundary heights of the input. This discrepancy is below 0.3 mm, a quarter of the 1.20 mm sheet thickness. The generation and reconstruction examples show how a fixed interface can support several candidate layouts in the shared design representation.

\begin{figure}[tp]
\centering
\includegraphics[width=\linewidth,height=205mm,keepaspectratio]{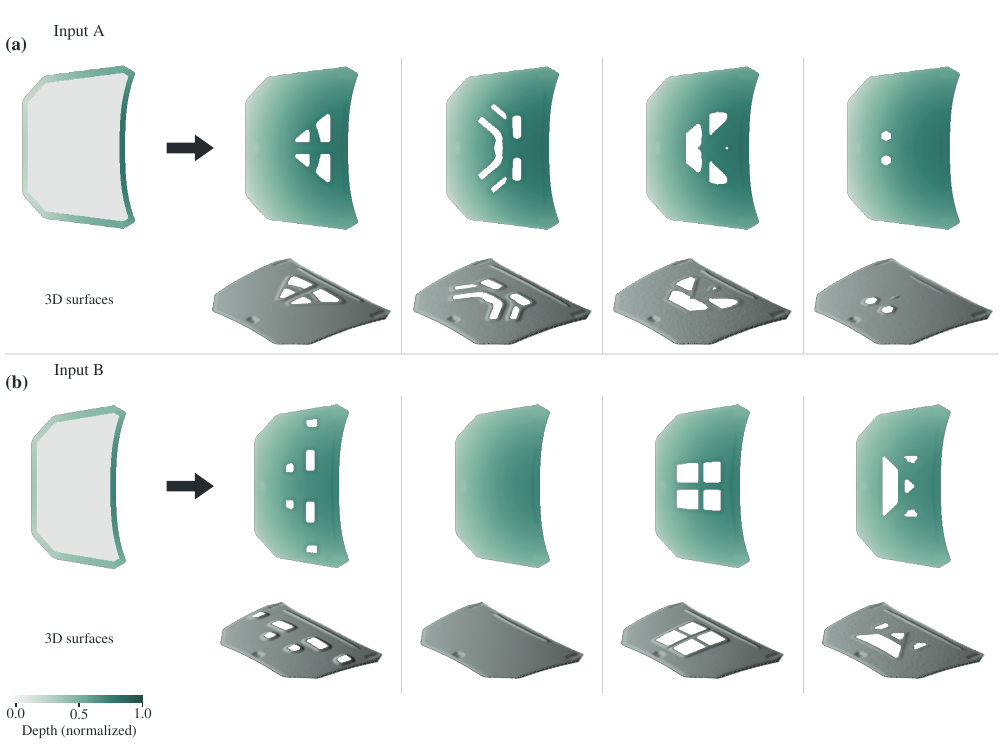}
\caption{Conditional generation under (a) Input A and (b) Input B, shown as depth maps and reconstructed surfaces from the first of three full-data training runs ($N$ = 7,660, R1). Inputs show the outer domain in grey and the prescribed boundary heights as a coloured band. For Input A, four of the eight saved samples from two dataset entries that share this input are shown. For Input B, the first four samples are shown. Symmetry is imposed by reflecting one half of each sample. Input boundary width is enlarged for visibility.}\label{fig:figure4}
\end{figure}

Training-data availability affects both surface quality and the variety of generated patterns (Fig. 5). Models trained with five or ten designs produce panels within the specified domain, but repeated opening patterns are common. The amount of training data therefore changes the variety and smoothness of the generated panels but not the encoded conditions. With five designs, the model from the second training run, R2, generates Input A layouts that differ from all five of its training designs and from one another (Supplementary Fig. S2). Its Input B samples and the samples from R1 and R3 mostly show a single layout. Five designs can therefore yield novel and varied layouts, although this depends on the training run and input. The three training runs vary both the training seed and the selected designs, so the differences between them do not isolate either source of variation.

\begin{figure}[tp]
\centering
\includegraphics[width=\linewidth,height=205mm,keepaspectratio]{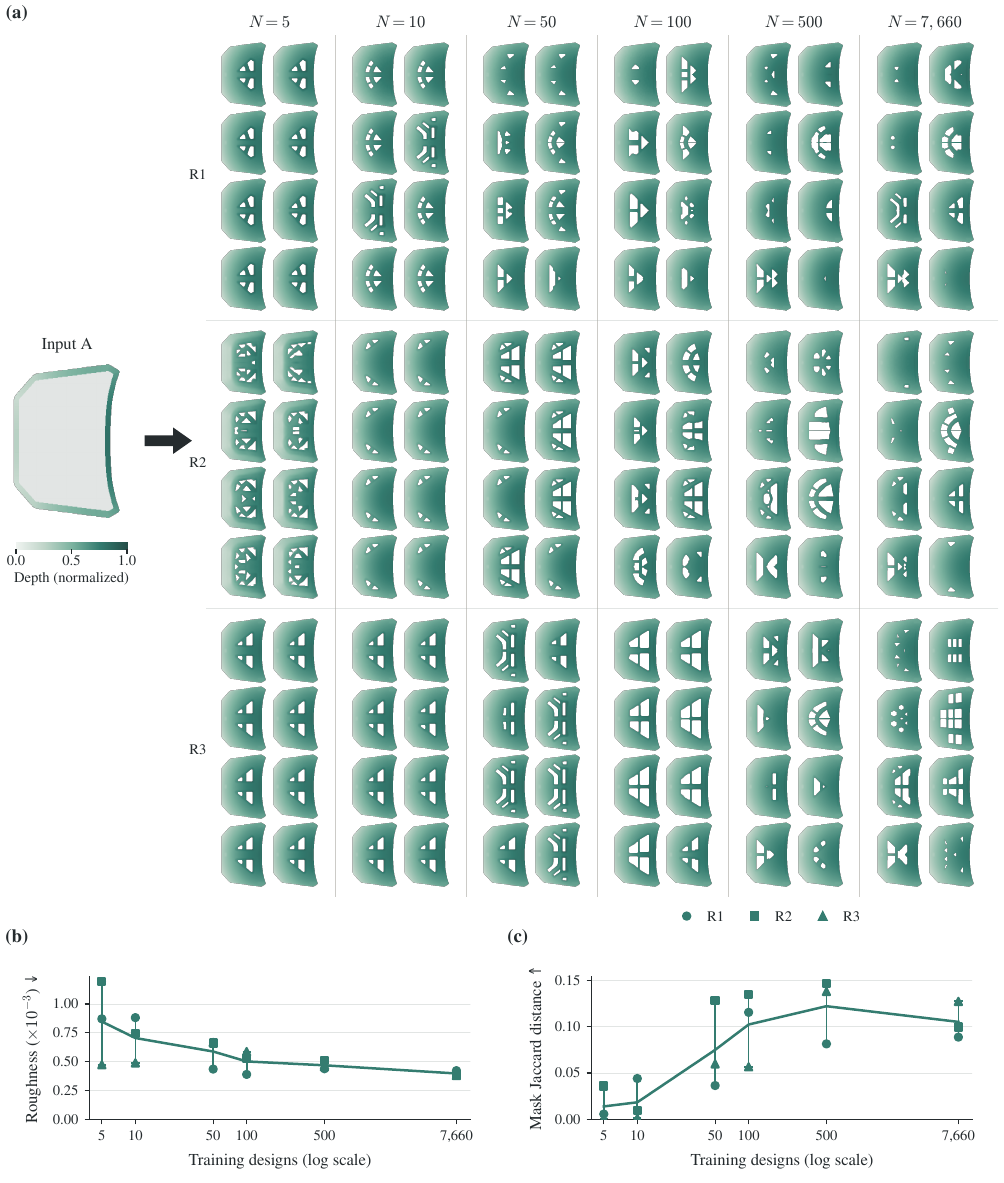}
\caption{Effect of training-data size $N$. Panels show (a) generated depth maps, (b) fine-scale depth fluctuations (roughness, $\downarrow$), and (c) differences between generated material masks (Jaccard distance, $\uparrow$). R1--R3 denote independent training runs, with eight samples each under Input A, the same input as in Fig. 4(a). Metrics are averaged over Inputs A and B with eight samples each. Lines and bars show the mean and range across runs. Input boundary width is enlarged for visibility.}\label{fig:figure5}
\end{figure}

Enlarging the training set generally reduces fine-scale height variation. Mean roughness decreases from $0.847\times10^{-3}$ at five designs to $0.397\times10^{-3}$ with all 7,660 designs. Pattern diversity, measured by mean pairwise mask Jaccard distance, rises from 0.014 at five designs to 0.122 at 500 designs and remains similar for the full-data model at 0.105, with overlapping ranges across training runs. A generator trained on 500 designs, 6.5\% of the training set, therefore already reaches the layout variety of the full-data model. The full-data model has the lowest roughness in this comparison. The same trends appear for one design from each of the ten validation families (Supplementary Table S4). These metrics describe complementary properties rather than a single ordering of design quality.

The comparison shows how the framework behaves as the available design collection grows. Several opening layouts can be learned without using the entire dataset, while very small datasets lead to greater variation between training runs and more repeated patterns. The displayed results use the same final training budget for all dataset sizes. Consequently, Fig. 5 measures the effect of data availability under that fixed procedure, rather than the best achievable result after separately tuning each data regime.

The five-design training-budget study shows a trade-off between layout variation and surface smoothness (Supplementary Figs. S3 and S4). Mean roughness falls from $6.900\times10^{-3}$ at 500 updates to $3.515\times10^{-3}$ at 10,000 updates, compared with $0.847\times10^{-3}$ for the original 240,000-update reference. Mean mask Jaccard distance reaches a peak of 0.107 at 2,500 updates and declines to 0.054 at 10,000 updates, while the reference gives 0.014. Shorter training retains more variation in the material masks, but at 2,500 updates or fewer the surfaces have not yet developed the height profile of the reference (Supplementary Fig. S3).

\subsection{Performance prediction and candidate ranking}
The depth representation supports performance assessment of designs from held-out design families (Fig. 6). Across 1,072 test designs, the three-model mean predicts deformation with an MAE of 0.6702 mm, an RMSE of 1.193 mm, and an $R^2$ of 0.809. Mass prediction achieves an MAE of 0.1885 kg, an RMSE of 0.2274 kg, and an $R^2$ of 0.983, supporting mass comparison across designs. These errors correspond to mean absolute percentage errors of 5.9\% for deformation and 1.4\% for mass. Mass computed from the official meshes in millimetres at a uniform steel density gives a mean absolute percentage error of 2.0\% on the same designs, because the labels carry family-dependent offsets from the geometric mass. The predictor therefore reproduces the label mass more closely than a geometric calculation from the original meshes. Table 2 lists the errors reported by earlier CarHoods10k predictors, whose test sets differ from ours. Two of them draw test designs from families that also appear in training, and one uses a six-surface subset. On families absent from training, STAMP reaches a deformation error close to the lowest reported values and a mass error below all reported values.

\begin{figure}[tp]
\centering
\includegraphics[width=\linewidth,height=205mm,keepaspectratio]{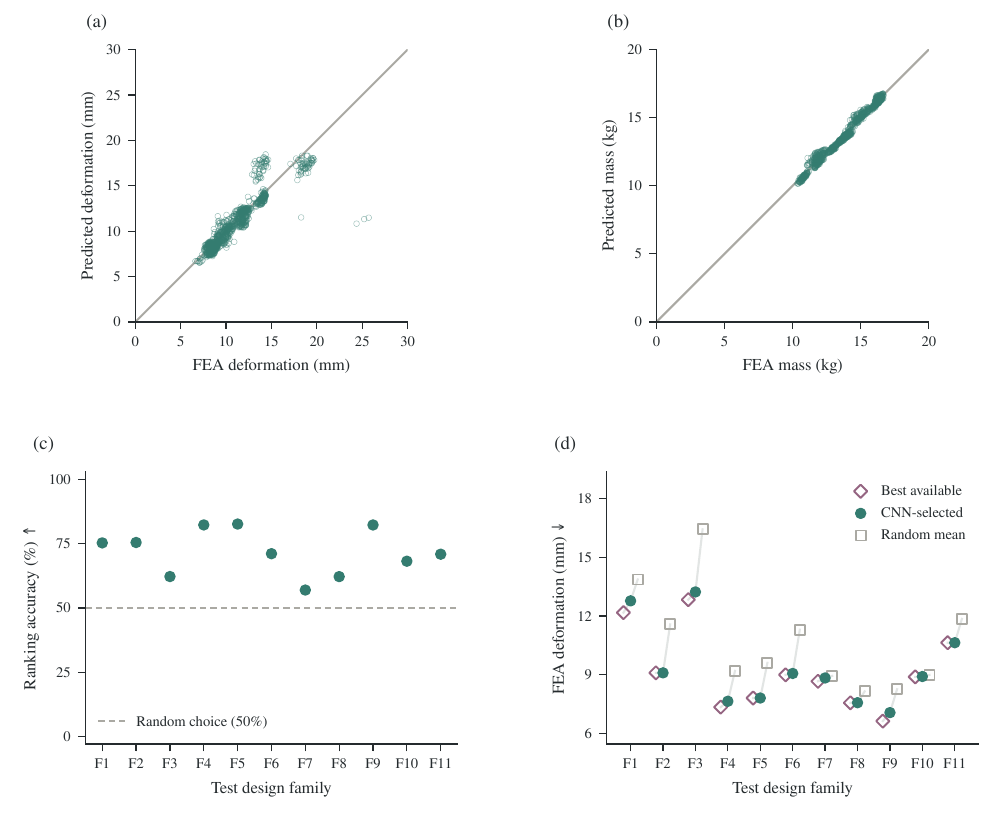}
\caption{Structural prediction on 1,072 test designs from 11 design families. (a, b) Deformation and mass predicted by the mean of the three CNNs versus FEA, with the diagonal marking equality. (c) Pairwise ranking accuracy. (d) FEA deformation of the best available and CNN-selected candidates, and the expectation for uniform random selection, within each family of 92--100 candidates.}\label{fig:figure6}
\end{figure}

The predictor also identifies promising designs within a design family. Each test family offers 92 to 100 dataset designs as candidates. Pairwise deformation ranking accuracy averages 71.8\% across the 11 test families, ranging from 57.0\% to 82.6\%. Selecting the candidate with the lowest predicted deformation gives an average FEA selection error of 0.180 mm, compared with an expected 1.591 mm for uniform random selection. The selected candidate has a smaller error than this random-choice expectation on every test family (11 of 11, two-sided sign test $p<0.001$). Predictor-based selection thus reduces the selection error by 89\% relative to the expected error of random choice. These results show how STAMP can guide candidate selection directly from its design representation.

Each network evaluates all 1,072 designs in 0.636--0.685 s with the inputs and model already loaded. Together, the prediction and selection results establish the performance-evaluation step that connects conditional generation to gradient-based shape optimization.

\begin{table}[tp]
\centering
\caption{Deformation and mass errors of CarHoods10k predictors. Values for earlier studies are those reported in the cited papers on their own test sets. MAPE is the mean absolute percentage error.}\label{tab:table2}
\begin{tabularx}{\linewidth}{p{34mm}p{36mm}Yrr}
\toprule
Study & Input & Test designs & \shortstack{Deformation\\MAPE (\%)} & \shortstack{Mass\\MAPE (\%)} \\
\midrule
Wollstadt et al. \cite{carhoods2022} & Point-cloud latent code & Random 10\% of all designs & 5.6--8.1 & 1.9--3.8 \\
Indupally et al. \cite{indupally2026afg} & Attributed feature graph & 301 designs distinct from training, six surfaces & 10.6 & 3.0 \\
Chavare \cite{chavare2026} & Point cloud & 20\% of each family & 5.7 & 2.0 \\
STAMP & Depth map & 11 families absent from training & 5.9 & 1.4 \\
\bottomrule
\end{tabularx}
\end{table}

\subsection{Performance-driven shape optimization}
The manufacturing-informed design space enables STAMP to optimize both generated alternatives and existing panels while preserving the drawing-direction condition on the reference surface, the nominal thickness, and the prescribed fixed regions (Fig. 7). The generated panel from Fig. 1 illustrates how optimization changes its surface profile while retaining its outline, openings, and designated fixed regions. The depth-difference map shows these changes as fractions of the depth range, and the aligned section shows them with an exaggerated vertical axis. By construction, the parameterization leaves the fixed boundary heights, the designated hinge and latch regions, and the opening masks unchanged and keeps every result within its depth-change bound. All 24 searches confirm these properties.

\begin{figure}[tp]
\centering
\includegraphics[width=\linewidth,height=205mm,keepaspectratio]{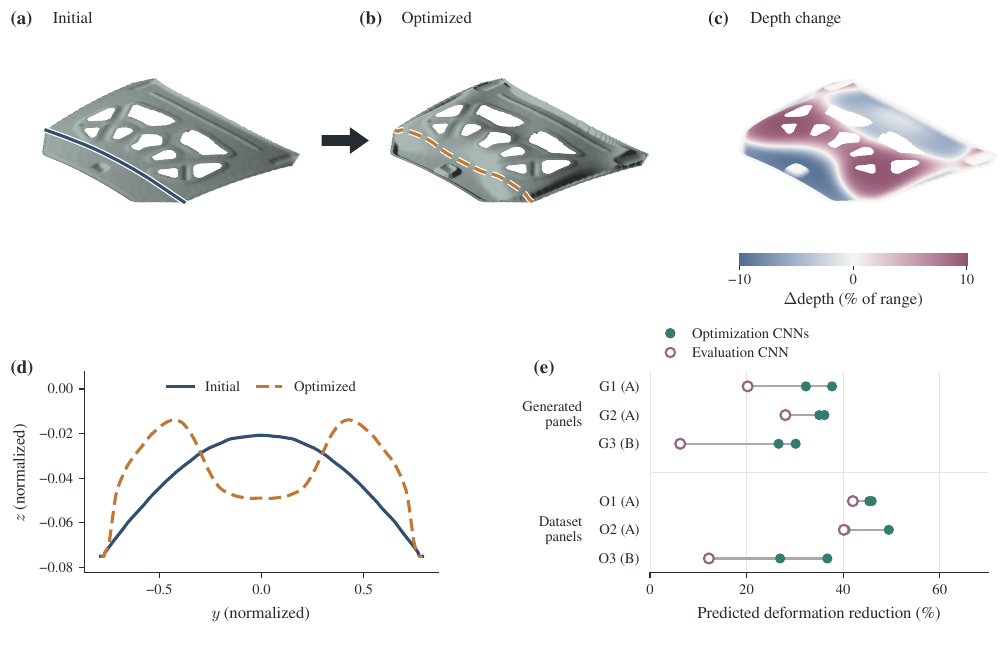}
\caption{Optimization with a 10\% depth-change bound. (a, b) Initial and optimized geometry at actual scale, (c) depth change, and (d) section profiles along the line marked in (a) and (b), for the generated panel in Fig. 1. Section heights are in normalized OBJ coordinates, with the vertical axis exaggerated for visibility. (e) Predicted deformation reductions for G1--G3 and O1--O3, with filled markers for the two optimization CNNs and open markers for the evaluation CNN. Letters A and B in parentheses identify the inputs.}\label{fig:figure7}
\end{figure}

Increasing the available depth change gives the optimizer more freedom to reduce predicted deformation. Averaged across the six starting designs, the evaluation CNN predicts reductions of 15.0\%, 24.8\%, 31.8\%, and 35.2\% at the four successive bounds (Table 3). On average, each further 5\% of allowed depth change adds less reduction than the previous step, whereas the area-based mass increase keeps growing from 1.3\% to 6.0\%. The two optimization CNNs and the evaluation CNN predict a reduction for every optimized design. The 10\% bound illustrated in Fig. 7 is one of these experimental settings. For Input A, it corresponds to $\pm$31.0 mm in surface height, and the four bounds span $\pm$15.5 to $\pm$62.0 mm, before the local taper and depth envelope further restrict the permitted motion. At this bound, the mean of the optimization CNNs predicts that the deformation of G1 falls from 12.60 to 8.19 mm, a reduction of 35.0\%. The evaluation CNN predicts a 20.2\% reduction for the same shape. Its surface area, and thus its mass at the prescribed nominal thickness, increases by 2.7\%.

The improvement and mass trade-off depend on the starting panel. At the 10\% bound, the deformation reductions predicted by the evaluation CNN range from 6.3\% to 42.0\%, with mass increases of 2.7\% to 4.0\%. G3 shows the largest difference in deformation reduction between the optimization and evaluation models. The two optimization CNNs predict 26.6--30.2\%, while the evaluation CNN predicts 6.3\%. Reporting all six cases distinguishes agreement on the direction of improvement from agreement on its magnitude.

Each 200-update search, including the evaluations by the third CNN, takes 1.29--1.44 s with the models already loaded. Together, these results demonstrate a direct path from conditional panel generation to rapid performance-driven shape optimization. The representation retains the prescribed geometric design conditions throughout the search, and the computed mass changes make the associated trade-offs visible.

\begin{table}[tp]
\centering
\caption{Optimization results, reported as mean (range) across six starting panels. Deformation reductions are predicted. Optimization uses two CNNs, and a third CNN evaluates deformation independently of the search. Mass change is computed from the surface area at a uniform nominal thickness.}\label{tab:table3}
\begin{tabularx}{\linewidth}{p{22mm}YYY}
\toprule
\shortstack[l]{Depth-change\\bound (\%)} & Deformation\newline reduction,\newline optimization CNNs (\%) & Deformation\newline reduction,\newline evaluation CNN (\%) & Area-based\newline mass change (\%) \\
\midrule
5 & 25.5 (19.4--32.0) & 15.0 (5.8--23.9) & +1.3 (+1.0--+1.9) \\
10 & 37.0 (28.4--45.7) & 24.8 (6.3--42.0) & +3.2 (+2.7--+4.0) \\
15 & 43.2 (32.1--51.9) & 31.8 (7.5--49.8) & +4.6 (+4.2--+5.2) \\
20 & 46.0 (33.0--55.1) & 35.2 (8.6--52.3) & +6.0 (+5.1--+6.5) \\
\bottomrule
\end{tabularx}
\end{table}

\FloatBarrier
\section{Discussion}

\subsection{What a manufacturing-informed representation enables}
STAMP defines a geometric design space for formed sheet panels through surface height, material region, and prescribed nominal thickness. Within the single-valued reference-surface class, these variables accommodate curved profiles, ribs, and openings without restricting the design to a predefined set of feature templates. The drawing-direction condition of the reference surface and the prescribed nominal thickness remain satisfied as layouts are generated and surfaces are modified. Structural objectives can therefore guide surface changes without manufacturing penalty weights for these conditions. This formulation makes the permitted design freedom explicit and keeps the geometric manufacturing requirements independent of the structural objective.

For engineers, this design space separates the interface to retain from the interior geometry to explore. Conditional generation proposes different material and opening layouts within a prescribed outer domain. Shape optimization develops a chosen layout through surface-height changes while retaining its outline, openings, and designated fixed regions. Engineers select the depth-change bound within the permitted height range. The allowed depth change and computed mass change then let engineers assess how much predicted structural benefit accompanies further geometric modification. In this study, the predicted benefit of each further increase in the bound declines on average, while each increase still adds mass (Table 3). Manufacturing knowledge thus governs both the alternatives that can be created and the changes available to develop them.

Data availability and training budget influence different aspects of the generated designs. A generator trained on 500 designs, 6.5\% of the training set, already reaches the layout variety of the full-data model, and additional data mainly smooths the generated surfaces. From only five designs, one of three training runs yields novel and varied layouts. Because the depth map is an image, STAMP can use convolutional generators, whose locality and translation equivariance let them compose new layouts from local patterns of the training designs \cite{kamb2024creativity}. Engineering design collections are often small \cite{hong2025deepjeb,kwon2025limited}, so reaching the full layout variety with a few hundred designs widens the range of problems that STAMP can address. Layout variety helps engineers explore alternatives, while surface smoothness affects the geometry available for subsequent evaluation and modification.

Performance prediction makes the generated alternatives useful for design decisions by supporting candidate selection and directing surface changes. A mass constraint or a multiobjective formulation could incorporate an explicit preference on mass into the search.

In vehicle development, an inner panel must also keep clearance from the outer panel above it and to the components beneath it. Because each value in the depth map is a height at a fixed planar location, these packaging requirements reduce to lower and upper limits on each value, the same form as the depth-change bound used here. Such envelopes were not evaluated because the public dataset contains no outer-panel or package geometry.

\subsection{Geometric scope and application requirements}
The geometric scope of STAMP is defined by the drawing direction and the single-valued reference-surface condition, independently of the component name or training dataset. Sheet panels that satisfy this condition already fall within the representation, including surfaces with different outer shapes, ribs, and opening patterns. This scope includes qualifying panels made by stamping and some parts made by hydroforming, rubber-pad forming, stretch forming, or incremental forming.

Applying the design framework to another component requires specifying its drawing direction, nominal thickness, geometric interfaces, and design conditions. The generator requires suitable geometries for the intended component family, and performance predictors require analysis data for the relevant materials, loads, and supports. Training or adapting these models and evaluating them for the new application are separate from establishing which reference surfaces the representation can describe.

Thermoformed polymer sheets and formed composite sheets can also share the reference-surface description when their reference surfaces satisfy the same single-valued condition. Their constitutive behaviour, thickness changes, and process limits require application-specific treatment. Material identity, thickness, or laminate orientation could enter as additional fields or fixed conditions. A bonded laminate need not introduce a separate geometric surface for every ply when one reference surface and layer properties describe its configuration. Additional fields would extend the present formulation and would need validation for each new part and process.

The representation could be extended from individual panels to multi-panel structures by assigning a separate depth-map channel to each constituent panel. Each channel would encode its panel's reference-surface height and material region, with nominal thickness prescribed separately for each panel. For panels that are individually single-valued along a shared drawing direction, the channels would distinguish surfaces that overlap at the same planar position. This would provide a component-wise geometric description of layered or spatially separated sheet-panel assemblies, such as the inner and outer panels of an automotive hood. Relative placement, connections, clearances, and interference would require explicit treatment. Assembly-level generation, structural prediction, and coupled optimization would require suitable data and validation beyond the present single-panel experiments.

\subsection{Limitations and validation scope}
The present implementation describes one panel with one depth map. Surfaces with several heights at one projected position, or walls parallel to the drawing direction, cannot be represented exactly by one field. Finite resolution introduces a separate limitation. Steep surfaces and small features may lose detail even when the reference surface is single-valued. The reconstruction assessment reports the resulting geometric loss, and the supplementary material gives detailed measurements. A prescribed nominal thickness defines the offset of the design surface, not the thickness distribution after forming.

We evaluate this general formulation on one public hood design problem. Generation comparisons use two geometric inputs and three training runs per dataset size, while predictor testing spans 11 held-out design families. These experiments characterize behaviour within the evaluated problem and do not establish transfer to another component or loading condition. The varying within-family ranking accuracy also shows why aggregate prediction error alone is insufficient for design selection.

Optimization results are predicted responses and have not been confirmed with the finite-element setup that produced the dataset labels. The evaluation CNN supplies no gradients and does not choose the final state, but it shares the training data and architecture of the optimization CNNs. Its agreement with the optimization CNNs on improvement therefore complements the search results without providing a physical error bound. CAD reconstruction, structural analysis, and forming assessment of retained candidates are needed before engineering acceptance. Such assessment must address thinning, wrinkling, tearing, springback, and assembly compatibility, since preserving the geometric design conditions alone does not guarantee manufacturability.

\section{Conclusion}

STAMP introduces a directly editable sheet-panel representation that satisfies the drawing-direction condition on the reference surface and the prescribed nominal thickness by construction. These geometric manufacturing requirements remain satisfied, without manufacturing penalty terms, throughout conditional generation and gradient-based shape optimization.

Experiments on CarHoods10k demonstrate alternative layouts under prescribed boundaries, full layout variety from 500 training designs, useful candidate selection among existing designs, and predicted deformation reductions with accompanying mass changes. Engineers can therefore explore layouts for a prescribed interface and develop a selected panel through controlled surface changes within the manufacturing-informed design space.

\section*{Acknowledgements}

This work was supported by grants from the Ministry of Science and ICT (GTL24033-000, N1026002, and No. 2022-0-00986), the Ministry of Trade, Industry and Energy (RS-2025-02317327 and RS-2025-25444634), the Ministry of Oceans and Fisheries (PET0050), and Korea Hydro \& Nuclear Power Co., Ltd. (No. 8-Tech-07).

\section*{CRediT authorship contribution statement}

Jihoon Kim: Conceptualization, Methodology, Software, Validation, Formal analysis, Investigation, Data curation, Visualization, Project administration, Writing -- original draft, Writing -- review \& editing. Jun Hyeok Song: Software, Writing -- review \& editing. Wonho Ha: Software, Writing -- review \& editing. Namwoo Kang: Supervision, Funding acquisition, Writing -- review \& editing.

\clearpage
\renewcommand{\shorttitle}{STAMP: Supplementary material}
\setcounter{figure}{0}
\setcounter{table}{0}
\renewcommand{\thefigure}{S\arabic{figure}}
\renewcommand{\thetable}{S\arabic{table}}
\renewcommand{\theHfigure}{S\arabic{figure}}
\renewcommand{\theHtable}{S\arabic{table}}
\begin{center}
{\large\bfseries Supplementary material}
\end{center}

\section*{S1 Representation assessment}
We assess the reference-surface approximation using one fixed design from each of the 104 dataset-defined families. Visible reference points are sampled on a 512 $\times$ 512 grid in the common frame. For each candidate resolution (64, 128, and 256), distances are measured from these points to reconstructed surface triangles. The candidate fields are stored at the same numerical precision as the training inputs. Each design contributes one mean distance and one 95th-percentile distance. All 104 designs remain in the distributions, including those with larger errors.

At 256 $\times$ 256, the median mean and 95th-percentile distances are $0.142\times10^{-3}$ and $0.753\times10^{-3}$, respectively. At 128 $\times$ 128 they are $0.410\times10^{-3}$ and $2.076\times10^{-3}$. At 64 $\times$ 64 they are $1.247\times10^{-3}$ and $6.466\times10^{-3}$. Using each design's own scale, the 256 $\times$ 256 values correspond to 0.13 and 0.68 mm. The 512 $\times$ 512 sampling is a geometric reference, not a trained generator resolution.

We also compared the depth map with two representations that store about the same number of values, using the same 104 designs and reference points. The voxel grid divides the common frame box into 60 $\times$ 78 $\times$ 14 cubic cells of width 0.024, records whether each cell contains a reference point, and recovers the surface by marching cubes. The point cloud holds 21,845 points, or 65,535 coordinates, sampled uniformly by area on the reference surface. Its surface is the Delaunay triangulation of the points in the projection plane, after removing triangles whose longest edge exceeds $K$ times the mean point spacing. Small values of $K$ leave gaps in the material, and large values close the openings. We therefore evaluated $K = 2$, 3, 4, 6, and 8 and report, for each design, the value most favourable to the point cloud. Distances are measured from the reference points to each reconstructed surface. Within the filled outline of the reference, the opening region is the area without reference material, and the opening intersection-over-union compares this region with the area that each representation leaves open. Table S3 lists the median values. A signed-distance grid of the same size was also considered. Its node spacing is 17 to 20 times the sheet thickness of the meshes, so only 117 to 210 of the 65,520 nodes lie inside the sheet and its zero level set does not recover the panel. Learned implicit representations, which can be evaluated at any resolution, were not compared.
\section*{S2 Five-design training subsets}
Figure S2 shows every training design used in the three five-design training runs. R1--R3 identify paired subsets and training seeds, rather than noise samples from one model. The subsets and seeds remain the same in the additional training-budget study. The different runs therefore vary both training examples and initialization. The comparison does not isolate the effect of either factor alone.
\section*{S3 Training-budget evaluation}
A supplementary training-budget study used five designs per model, retaining the three paired subsets and training seeds of the fixed-budget comparison. Three new models were trained from scratch and evaluated at preset milestones of 500, 1,000, 2,500, 5,000, and 10,000 optimizer updates. The original 240,000-update models provided the reference. Each saved model state yielded eight samples for each of the same two geometric inputs, with the same seeds, smoothing, and reflection rule as Fig. 5. We averaged surface roughness over samples within each input, computed mask diversity from all 28 sample pairs, and weighted the two inputs equally. The plots show the means of the three training runs and their observed range. No checkpoint was selected using these results.

All new runs used an RTX PRO 6000 Blackwell GPU, while the original R2 reference used an A100. The comparison is a separate training-budget study, rather than a continuation of the original training trajectories. Figs. S3 and S4 show fixed Input A examples and the quantitative comparison, respectively. Across the 18 model states, the evaluation includes 288 samples and 1,008 within-input pairs. Recomputing the metrics for the original three $N = 5$ models reproduces the main data-size evaluation exactly.
\section*{S4 Optimization cases}
Table S1 reports all six starting panels at each of the four depth-change bounds selected for the experimental comparison. G1, G2, and G3 are generated panels. O1--O3 are dataset panels. Input A is shared by G1, G2, O1, and O2. Input B is used by G3 and O3. Optimization uses the mean deformation prediction of the two optimization CNNs. The evaluation CNN assesses the retained geometry without contributing gradients or selecting states. The table reports predicted deformation reductions and area-based mass changes. It does not report new FEA of optimized shapes.
\section*{S5 Training and reproducibility notes}
The original generation comparison used a common schedule of 1,000 cycles, each presenting 7,660 training examples in batches of 32, for 240,000 scheduled optimizer updates. At smaller dataset sizes, training examples were revisited to retain that budget. Seventeen run histories record the final update counter directly. The full-data R1 history records 1,000 completed epochs, equivalent to the 1,000 cycles, without a separate update counter. Table S2 reports the scheduled budget for all runs and the measured training and validation times from their histories. The additional five-design runs stop at 10,000 updates. Training and validation take 1.29--1.34 h per run, and milestone sampling takes a further 86--87 s.

The three predictor training runs use the same 7,660 training designs and 200-epoch schedule (24,000 updates per run). Each takes 38.1--38.2 min on an RTX PRO 6000. Inference times reported in the main text are measured separately on an A100 with the networks already loaded. Optimization timing also excludes model loading and file export. These measurements reflect the recorded hardware and what each timing includes. They do not represent equal hardware cost across the generation comparison.

The samples in Fig. 5 and Figs. S3--S4 were generated from the saved checkpoints on one A100 GPU. Sampling on another GPU type can change individual samples.
\section*{S6 Implementation details}
The generator uses a two-dimensional U-Net with three levels of 32, 64, and 128 channels and two layers per block. Training uses 1,000 diffusion timesteps.

The predictor network contains four convolutional blocks with 32, 64, 128, and 192 channels. Each block applies two 3 $\times$ 3 convolutions with batch normalization and SiLU activation, followed by average pooling. Adaptive pooling retains a 2 $\times$ 2 spatial layout. A fully connected layer with 256 features, SiLU activation, and dropout of 0.15 precedes the three outputs. Stress and deformation are transformed logarithmically, and all targets are standardized using training statistics. Predictions are converted back to physical units before ensemble averaging. Stress is trained jointly but is not used for selection or optimization. For the 1,072 test designs, the three-model mean predicts it with an MAE of 14.27 MPa, an RMSE of 18.56 MPa, and an $R^2$ of 0.537.

In the depth increment of Section 3.5, $F$ is the symmetrized B-spline field, $r$ bounds the depth change relative to the common depth range, and $E$ combines the movable region, boundary taper, and depth envelope. The movable region excludes the prescribed fixed pixels and the background. A three-pixel boundary band remains fixed, while material around internal openings can move with the surrounding surface. The latch and the two hinge mounting regions of each surface group are identified from flat regions in its reference depth map and dilated by three pixels. These regions remain fixed at their initial heights during optimization. A quintic smoothstep taper rises to one over a distance of eight pixels from fixed material. The depth envelope is 4$d_0$(1$-$$d_0$). We use the smaller envelope and taper value at each reflected pair and symmetrize the movable region, so the increment is symmetric and remains within the depth range. Existing asymmetry in the initial surface is retained. These are geometric design constraints, separate from the supports and loads used to generate the FEA labels.

Once the inputs are loaded, each network evaluates all 1,072 designs in 0.636--0.685 s on an NVIDIA A100 using batches of 64. This measurement includes input transfer and output collection. Optimization runs use an NVIDIA A100 80 GB GPU. Per-design timing includes the search itself, evaluation of all three CNNs at each state, gradient computation, and in-memory recording of results. It excludes model loading and file export.
\section*{S7 Validation-family generation check}
The data-size comparison in the main text uses two symmetry-compatible inputs. As a broader check, the same final models generated four samples for one design from each of the ten validation families, without reflection. Two of these families share the same boundary input, so the check covers nine distinct inputs. Roughness and mask Jaccard distance are computed on the decoded fields as in the main text and averaged over the ten cases. Table S4 reports the mean and range across the three training runs. As in the main comparison, roughness is lowest for the full-data model, and diversity is low with five or ten designs, rises up to 500 designs, and remains similar with all 7,660 designs.
\setlength{\tabcolsep}{4pt}
\begin{table}[tp]
\centering
\caption{Optimization results for all starting panels and depth
bounds.
Deformation reductions are predicted by the optimization CNN mean (Opt.) and the evaluation CNN (Eval.).
Mass change is computed from the surface area at a uniform nominal thickness.}
\begin{tabular}{rlrrr}
\toprule
\shortstack{Depth-change\\bound (\%)} & Panel &
\shortstack{Opt. deformation\\reduction (\%)} &
\shortstack{Eval. deformation\\reduction (\%)} &
\shortstack{Mass\\change (\%)}\\
\midrule
5 & G1 & 27.2 & 15.1 & +1.01 \\
5 & G2 & 25.7 & 18.4 & +1.16 \\
5 & G3 & 19.7 & 5.8 & +1.57 \\
5 & O1 & 32.0 & 23.9 & +1.11 \\
5 & O2 & 28.9 & 19.1 & +1.31 \\
5 & O3 & 19.4 & 7.9 & +1.92 \\
10 & G1 & 35.0 & 20.2 & +2.70 \\
10 & G2 & 35.6 & 28.0 & +3.06 \\
10 & G3 & 28.4 & 6.3 & +3.15 \\
10 & O1 & 45.7 & 42.0 & +2.80 \\
10 & O2 & 45.2 & 40.1 & +3.28 \\
10 & O3 & 32.2 & 12.2 & +3.96 \\
15 & G1 & 42.6 & 31.7 & +4.19 \\
15 & G2 & 43.8 & 35.2 & +4.57 \\
15 & G3 & 32.1 & 7.5 & +4.38 \\
15 & O1 & 51.9 & 49.8 & +4.40 \\
15 & O2 & 51.5 & 46.6 & +4.90 \\
15 & O3 & 37.2 & 20.0 & +5.24 \\
20 & G1 & 46.8 & 37.1 & +5.92 \\
20 & G2 & 47.8 & 39.9 & +6.03 \\
20 & G3 & 33.0 & 8.6 & +5.05 \\
20 & O1 & 55.1 & 52.3 & +6.11 \\
20 & O2 & 54.4 & 49.4 & +6.48 \\
20 & O3 & 39.1 & 23.9 & +6.29 \\
\bottomrule
\end{tabular}
\end{table}
\begin{table}[tp]
\centering
\caption{Training costs per run for the main experiments. Generator times include validation. Predictor times are elapsed training-run times. Updates denote the scheduled budget. Sampling and reporting are excluded.}\label{tab:tableS2}
\begin{tabularx}{\linewidth}{Xrrrr}
\toprule
Model and GPU & \shortstack{Training\\designs} & Runs & \shortstack{Updates\\per run} & Time (h) \\
\midrule
Generator, RTX PRO 6000 & 5--7,660 & 12 & 240,000 & 31.08--32.76 \\
Generator, A100 80 GB & 5--7,660 & 6 & 240,000 & 40.82--41.48 \\
Predictor, RTX PRO 6000 & 7,660 & 3 & 24,000 & 0.635--0.637 \\
\bottomrule
\end{tabularx}
\end{table}\begin{table}[tp]
\centering
\caption{Shape representations that store about 65,536 values, evaluated on the 104 designs of Section S1. Values are medians over designs, and distances are in common normalized coordinates. For the point cloud, the edge-length limit $K$ is chosen for each design to give the smallest mean distance or the largest opening IoU.}\label{tab:tableS3}
\begin{tabularx}{\linewidth}{Xrrrr}
\toprule
Representation & \shortstack{Stored\\values} & \shortstack{Mean distance\\($\times10^{-3}$)} & \shortstack{95th-percentile\\distance ($\times10^{-3}$)} & \shortstack{Opening\\IoU} \\
\midrule
Depth map, 256 $\times$ 256 & 65,536 & 0.142 & 0.753 & 0.94 \\
Point cloud, 21,845 points & 65,535 & 0.210 & 1.37 & 0.82 \\
Voxel grid, 60 $\times$ 78 $\times$ 14 & 65,520 & 9.51 & 16.8 & 0.70 \\
\bottomrule
\end{tabularx}
\end{table}\begin{table}[tp]
\centering
\caption{Generation metrics for one design from each of the ten validation families, with four samples each and no reflection. Values are the mean and, in parentheses, the range across three training runs.}\label{tab:tableS4}
\begin{tabular}{lrr}
\toprule
Training designs & Roughness ($\times10^{-3}$) & Mask Jaccard distance \\
\midrule
5 & 1.02 (0.99--1.06) & 0.035 (0.017--0.049) \\
10 & 0.80 (0.68--0.90) & 0.029 (0.021--0.038) \\
50 & 0.65 (0.65--0.66) & 0.087 (0.068--0.105) \\
100 & 0.65 (0.62--0.67) & 0.108 (0.098--0.114) \\
500 & 0.66 (0.60--0.71) & 0.141 (0.111--0.168) \\
7,660 & 0.55 (0.55--0.57) & 0.127 (0.106--0.148) \\
\bottomrule
\end{tabular}
\end{table}\begin{figure}[tp]
\centering
\includegraphics[width=\linewidth,height=205mm,keepaspectratio]{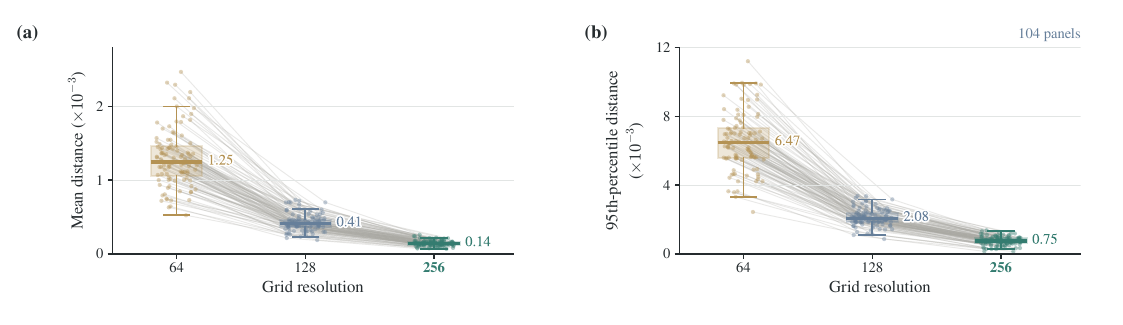}
\caption{Reconstruction distance across grid resolutions. (a) Mean and (b) 95th-percentile distances for 104 panels. Lines connect values for the same panel, dots show individual values, and boxes show quartiles and medians, with whiskers extending to 1.5 times the interquartile range. Distances use normalized coordinates and 512 $\times$ 512 visible-surface reference samples.}
\end{figure}
\begin{figure}[tp]
\centering
\includegraphics[width=\linewidth,height=205mm,keepaspectratio]{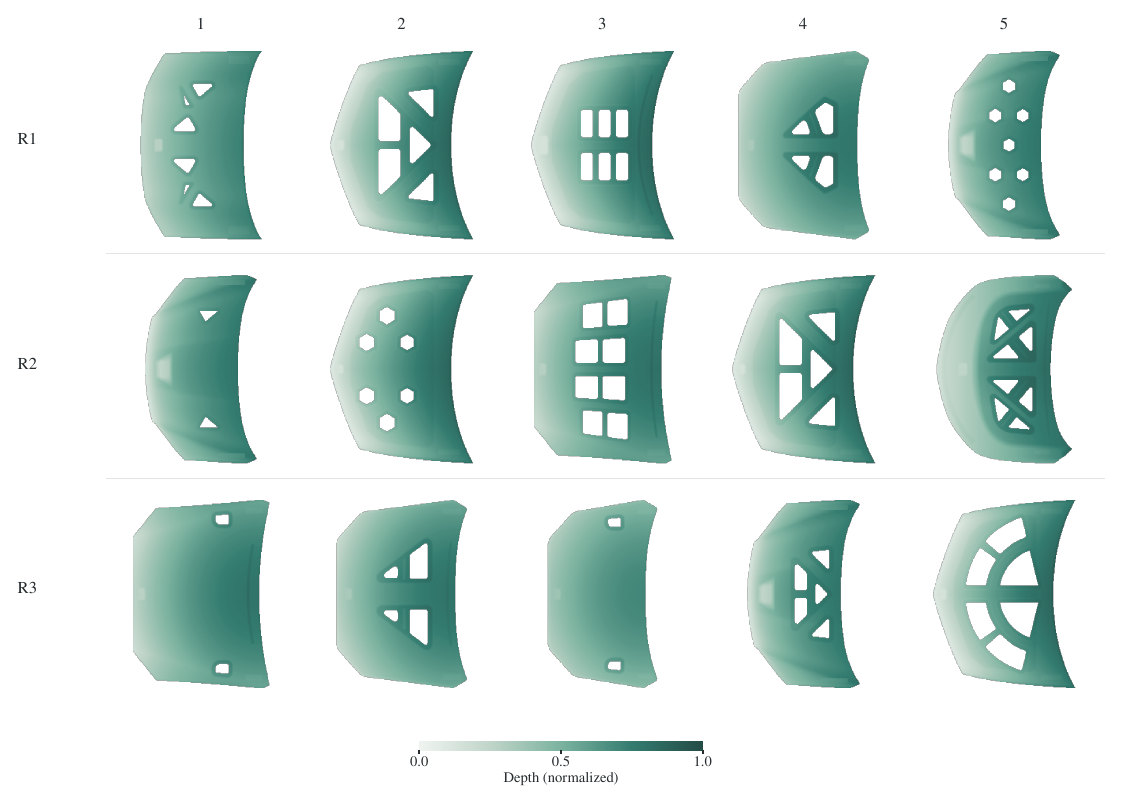}
\caption{All five training designs for each five-design training run (R1--R3). Columns follow the saved training subsets. Depth maps are shown as stored, without added symmetry or smoothing. All depth maps share the same colour scale.}
\end{figure}
\begin{figure}[tp]
\centering
\includegraphics[width=\linewidth,height=205mm,keepaspectratio]{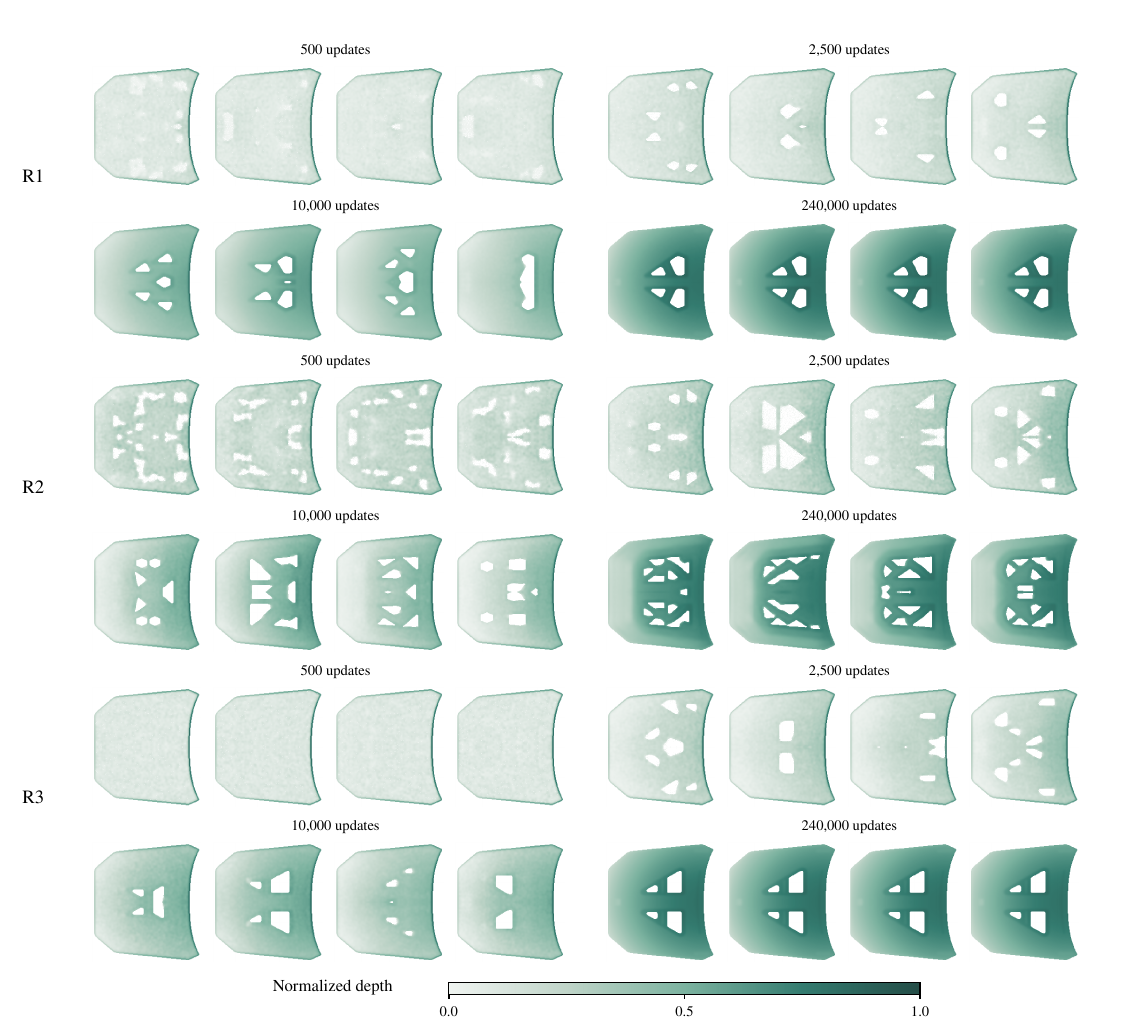}
\caption{Input A samples from the five-design models at four fixed training budgets. R1--R3 denote the three training runs, each shown with its first four noise samples per budget. Depth uses a common normalized 0--1 scale. Smoothing and reflection match Fig. 5. The 240,000-update samples come from the original models of Fig. 5.}
\end{figure}
\begin{figure}[tp]
\centering
\includegraphics[width=\linewidth,height=205mm,keepaspectratio]{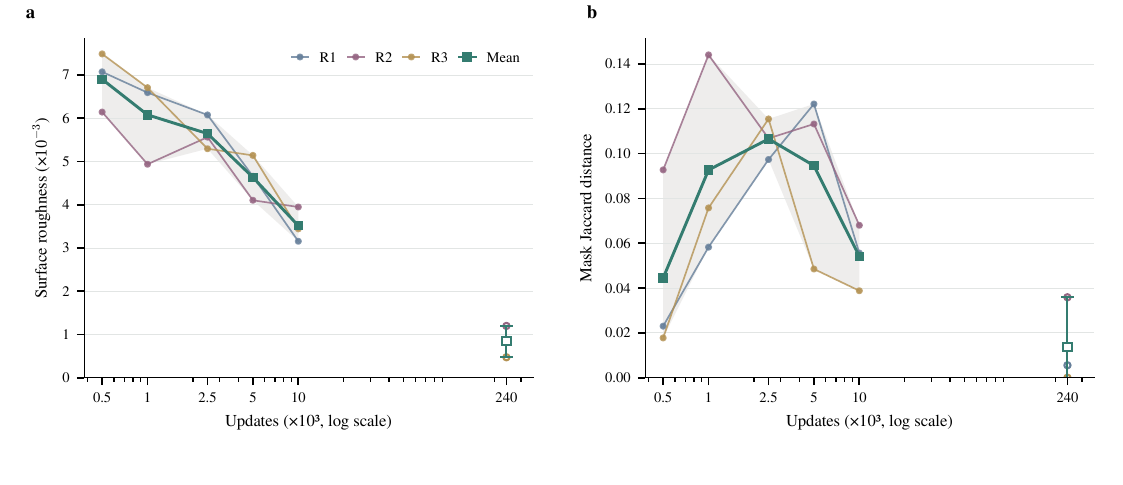}
\caption{Five-design training-budget comparison of (a) surface roughness and (b) mask diversity. Each training run weights the two inputs equally, using eight samples per input and all 28 pairs for diversity. Shading and reference whiskers show the observed range across the three runs. Open markers denote the original 240,000-update models. All curves use the same sampling, smoothing, and reflection as Fig. 5.}
\end{figure}

\end{document}